\documentclass[runningheads]{llncs}
\usepackage[T1]{fontenc}
\usepackage{graphicx}
\usepackage{xcolor}
\usepackage{hyperref}

\usepackage{tikz}
\definecolor{badgeblue}{HTML}{4285F4}
\definecolor{badgeyellow}{HTML}{F5A623}
\definecolor{badgedark}{HTML}{24292E}
\newcommand{\badge}[3]{
  \href{#1}{\tikz[baseline=(text.base)]{%
    \node[fill=#3, text=white, font=\footnotesize\bfseries\sffamily,
          inner xsep=5pt, inner ysep=2pt, rounded corners=2.5pt] (text) {#2};}}%
}
\usepackage{caption}  
\usepackage{wrapfig}
\usepackage{latexsym}
\usepackage[utf8]{inputenc}
\usepackage{inconsolata}

\usepackage{microtype}
\usepackage{booktabs}
\usepackage{xspace}
\usepackage{adjustbox}
\usepackage{amsmath}
\usepackage{multirow}
\usepackage{arydshln}
\usepackage{siunitx}
\usepackage{makecell}
\usepackage{array,makecell,graphicx}

\newcolumntype{C}[1]{>{\centering\arraybackslash}m{#1}}  
\newcolumntype{L}[1]{>{\raggedright\arraybackslash}m{#1}}

\newlength\savewidth

\newcommand{\dataset}{\textsc{ScratchMath}\xspace}

\usepackage{enumitem}
\usepackage{tcolorbox}
\newtcolorbox{AIBox}[2][]{
  enhanced,
  breakable,
  colframe=black, 
  colback=white,
  coltitle=white,
  fonttitle=\bfseries,
  attach boxed title to top left={xshift=1cm,yshift*=-\tcboxedtitleheight/2},
  boxed title style={
    colback=black,
    size=small,
    arc=1mm,
    boxrule=0pt,
    shadow={0.5mm}{-0.5mm}{0.5mm}{gray!60!black}
  },
  title=#2,
  arc=2mm,
  boxrule=0.6pt,
  #1
}

\begin{document}

\title{\textit{Can MLLMs Read Students' Minds?}\\Unpacking Multimodal Error Analysis in Handwritten Math}
\titlerunning{Unpacking Multimodal Error Analysis in Handwritten Math}

\author{Dingjie Song\inst{1} \and
Tianlong Xu\inst{2} \and
Yi-Fan Zhang\inst{4} \and
Hang Li\inst{5} \and
Zhiling Yan\inst{1} \and
Xing~Fan\inst{3} \and
Haoyang Li\inst{3} \and
Lichao Sun\inst{1} \and
Qingsong Wen\inst{2}\thanks{Corresponding author.}}
\authorrunning{D. Song et al.}
\institute{Lehigh University, Bethlehem, PA, USA\\
\email{\{dis724,zhy423,lis221\}@lehigh.edu} \and
Squirrel Ai Learning, Bellevue, WA, USA\\
\email{txu0915@gmail.com, qingsongedu@gmail.com} \and
Squirrel Ai Learning, Shanghai, China\\
\email{fanxing@songshuai.com, derek@squirrelai.com} \and
Institute of Automation, Chinese Academy of Sciences, Beijing, China\\
\email{yifanzhang.cs@gmail.com} \and
Michigan State University, East Lansing, MI, USA\\
\email{lihang4@msu.edu}\\[3pt]
\badge{https://bbsngg.github.io/ScratchMath/}{Project Page}{badgeblue}~%
\badge{https://huggingface.co/datasets/songdj/ScratchMath}{Dataset}{badgeyellow}~%
\badge{https://github.com/ai-for-edu/ScratchMath}{Code}{badgedark}}

\maketitle              

\vspace{-5mm}
\begin{abstract}
Assessing student handwritten scratchwork is crucial for personalized educational feedback but presents unique challenges due to diverse handwriting, complex layouts, and varied problem-solving approaches. Existing educational NLP primarily focuses on textual responses and neglects the complexity and multimodality inherent in authentic handwritten scratchwork. Current multimodal large language models (MLLMs) excel at visual reasoning but typically adopt an ``examinee perspective'', prioritizing generating correct answers rather than diagnosing student errors. To bridge these gaps, we introduce \textbf{\dataset}, a novel benchmark specifically designed for explaining and classifying errors in authentic handwritten mathematics scratchwork. Our dataset comprises 1,720 mathematics samples from Chinese primary and middle school students, supporting two key tasks: Error Cause Explanation (ECE) and Error Cause Classification (ECC), with seven defined error types. The dataset is meticulously annotated through rigorous human-machine collaborative approaches involving multiple stages of expert labeling, review, and verification. We systematically evaluate 16 leading MLLMs on \dataset, revealing significant performance gaps relative to human experts, especially in visual recognition and logical reasoning. Proprietary models notably outperform open-source models, with large reasoning models showing strong potential for error explanation. All evaluation data and frameworks are publicly available to facilitate further research.

\keywords{Multimodal Large Language Models \and Mathematical Reasoning \and Error Diagnosis \and Multimodal Systems \and Benchmarking \and Handwritten Recognition.}
\end{abstract}

\section{Introduction}

\begin{figure}[h]
    \centering
    \includegraphics[width=0.7\linewidth]{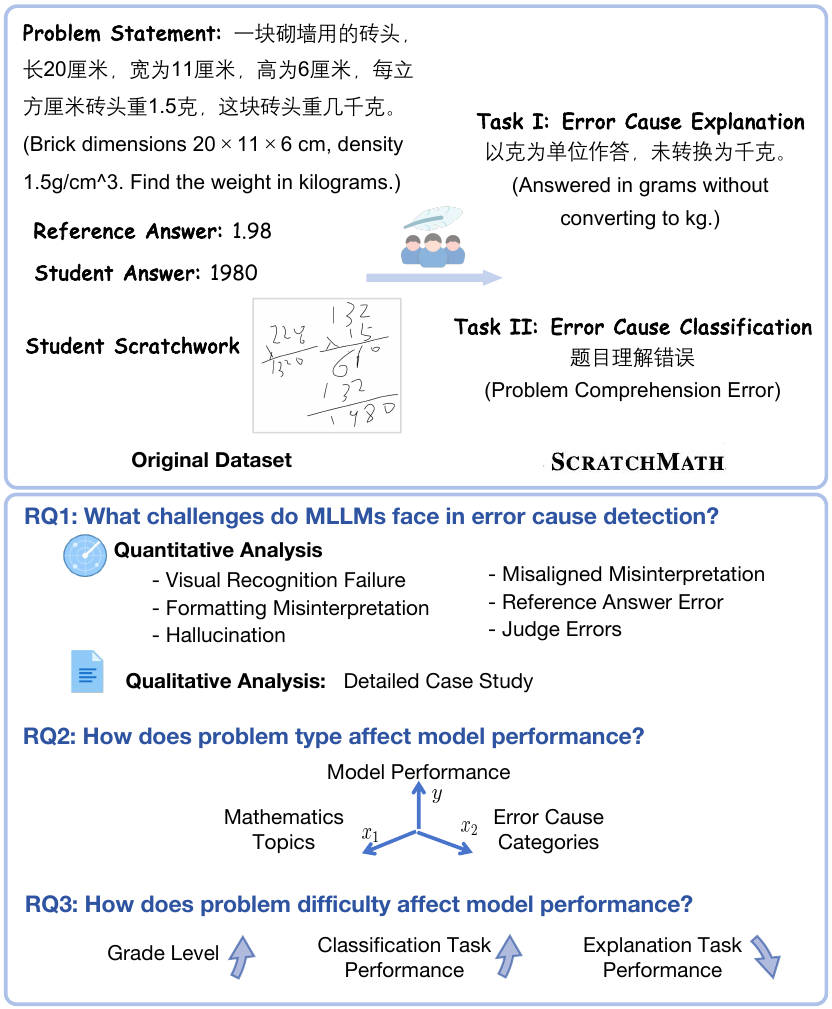}
    \caption{Overview of this work. (Top) An example illustrating two tasks proposed. (Bottom) Summary of the three research questions addressed in this study.}
    \label{fig:first}
\end{figure}

Automatically analyzing student work to provide precise, personalized feedback is critical in educational AI \cite{zhang2025correctness,kinder2025effects,davalos2025llms}. Teachers often diagnose misconceptions and errors by examining students' handwritten scratchwork \cite{caraeni2024grading}. Authentic scratchwork reflects individual cognitive processes but introduces unique challenges: ambiguity in symbol recognition (e.g., confusion between ``1,'' ``l,'' and ``|''), complex spatial layouts (e.g., fractions, superscripts), and personalized problem-solving strategies \cite{caraeni2024grading}. Accurate automated analysis of scratchwork can significantly enhance personalized teaching interventions \cite{yan2024errorradar}.

Previous educational NLP studies utilized rule-based systems or machine learning classifiers for error detection \cite{botelho2023leveraging,mcnichols2023algebra}, but these approaches lack generalizability and rely heavily on expert-defined error types. Recent work involving fine-grained LLM-based analyses using cognitive theory-guided strategies \cite{jiang2024pedcot} or iterative feedback loops \cite{chen2024boosting} mostly address textual answers, neglecting multimodal inputs such as handwritten scratchwork.

While multimodal large language models (MLLMs)~\cite{liu2023llava,bai2023qwen} excel at visual reasoning tasks, they primarily adopt an ``examinee perspective,'' focusing on generating correct answers rather than analyzing student solutions to diagnose errors—a perspective analogous to that of an educator or examiner \cite{zhang2024mathverse,wang2024measuring,yan2024errorradar}. Additionally, recent multimodal benchmarks, such as ErrorRadar \cite{yan2024errorradar}, and MathAgent \cite{xu2024mathagent}, often utilize structured data, limiting their effectiveness in capturing the complexity of authentic handwritten scratchwork and focusing mostly on error classification rather than detailed explanations.

To address these gaps, we introduce \textbf{\dataset}, a novel benchmark specifically designed for explaining and classifying errors in authentic handwritten scratchwork. Our dataset comprises 1,720 Chinese mathematics samples from primary and middle school students, covering five critical mathematical topics: Numbers and Expressions, Equations and Functions, Geometry and Measurement, Applied Mathematics, and Statistics and Probability. The dataset supports two essential tasks: \textbf{Error Cause Explanation (ECE)} and \textbf{Error Cause Classification (ECC)}. Based on student scratchwork, we define seven student error types, including \textit{Problem Comprehension Error} and \textit{Calculation Error}. The annotation process employs a human-machine collaborative approach, initially using MLLM for preliminary annotations, followed by multiple stages of expert labeling, review and verification to ensure accuracy and reliability.

We systematically evaluate 16 leading MLLMs (e.g., o4-mini~\cite{hurst2024gpt}, QVQ~\cite{qvq-72b-preview}) on \dataset with extensive analysis (see Figure~\ref{fig:first}). Results reveal significant gaps compared to human experts, particularly in correcting visual recognition errors and understanding logical transitions in multi-step solutions. Notably, proprietary models significantly outperform open-source models, and large reasoning models show promising capabilities, especially on the explanation task.
Our primary contributions are threefold:
\begin{enumerate}
    \item Introducing a novel multimodal error-detection and explanation benchmark task, specifically tailored for educational settings;
    \item Developing and publicly releasing the first high-quality, multimodal dataset of authentic student handwritten scratchwork, annotated via rigorous human-machine collaboration;
    \item Conducting the first evaluation of state-of-the-art MLLMs on this task, including detailed analyses highlighting their capabilities and limitations.
\end{enumerate}

\section{Related Work}

\label{sec:relatedwork}

\subsection{LLMs and MLLMs for Education}

Research on LLMs as AI tutors prioritizes pedagogical alignment and practical feedback~\cite{wang2026large}. For example, a LLaMA model was fine-tuned using GPT-4-based rubrics \cite{Lee2025TutorDPO}. Studies also demonstrate that adaptive LLM-generated feedback effectively boosts student motivation \cite{kinder2025effects}, and multimodal LLMs (MLLMs) can effectively summarize diverse learner data to aid teachers' assessments \cite{davalos2025llms}. However, existing MLLM-based grading methods for handwritten student solutions \cite{caraeni2024grading,Liu2024HandwrittenGrading} often struggle due to the complexity of authentic scratchwork. Despite advances in automatic scoring and feedback generation, few studies focus explicitly on pinpointing and explaining the precise reasoning failures within authentic handwritten scratchwork.

\subsection{LLMs and MLLMs for Mathematical Reasoning}

Beyond text-based mathematical reasoning, recent studies highlight challenges faced by MLLMs in interpreting diagrams, handwritten derivations, and visual reasoning tasks~\cite{yan2025survey}. Benchmarks such as MathVerse \cite{zhang2024mathverse}, MATH-V \cite{wang2024measuring}, and MileBench \cite{song2024milebench} reveal that even advanced models overlook crucial visual details. Specialized methods like Math-LLaVA \cite{Shi2024MathLLaVA} and LLaVA \cite{liu2023llava} have not fully resolved these issues. Recent multimodal benchmarks, including ErrorRadar \cite{yan2024errorradar} and MathAgent \cite{xu2024mathagent}, mainly use structured or semi-structured inputs, emphasizing error localization or classification rather than detailed explanations. Moreover, cognitive theory-guided approaches \cite{jiang2024pedcot} and iterative feedback strategies \cite{chen2024boosting} remain limited to text-only contexts.

Our work is also related to Handwritten Mathematical Expression Recognition (HMER), which converts handwritten mathematical notation into machine-readable formats. The CROHME competition series~\cite{mahdavi2019crohme} has served as the primary benchmark for this task, driving progress from structural approaches to neural encoder-decoder and transformer-based models~\cite{truong2024survey}. However, HMER focuses on symbol-level recognition accuracy, whereas \dataset targets a fundamentally different goal: diagnosing the \textit{reasoning errors} behind student solutions, which requires understanding both the visual content and the underlying mathematical logic. Our work addresses these complementary gaps by explicitly evaluating multimodal error detection and detailed explanation within authentic handwritten student scratchwork.

\section{The \dataset Benchmark}
\label{sec:dataset}

\begin{figure*}[t]
    \centering
    \includegraphics[width=1\linewidth]{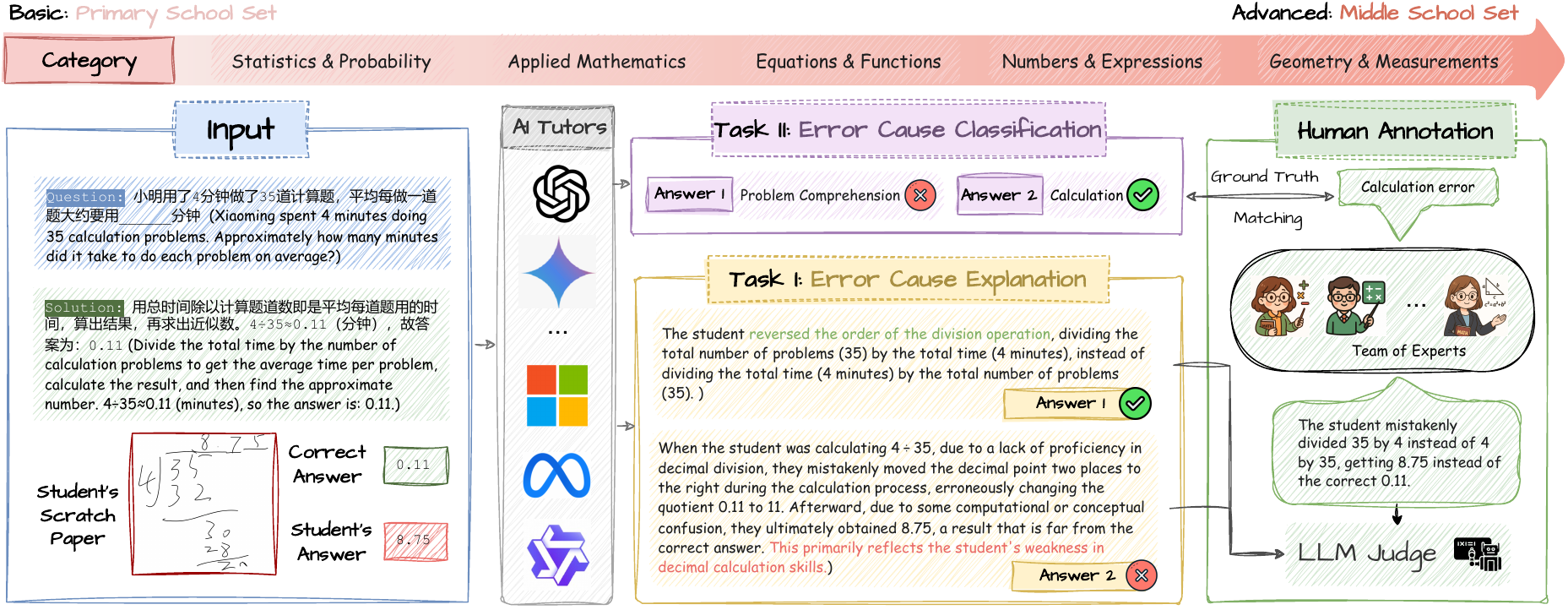}
    \caption{Overview of \dataset (with models' answer and human labeled answer translated in English). It illustrates dataset structure, tasks (ECE and ECC), multimodal model predictions, and expert human annotations.}
    \label{fig:overview}
\end{figure*}

\subsection{Task Definition and Taxonomy.}
\label{sec:dataset1}

Our goal is to evaluate MLLMs' ability to detect and explain errors in student solutions to math problems. A primary challenge is interpreting student scratchwork, which often combines diverse elements (e.g., handwritten text, symbolic notation, drawings) and requires integration with logical mathematical reasoning. Figure \ref{fig:overview} provides an overview of our task setup and evaluation framework.

\paragraph{Formal Task Definition}
Formally, each instance is defined by the following tuple:

$$
(Q, A_{\text{ref}}, S, A_{\text{stud}}, I)
$$

where $Q$ is the problem statement, $A_{\text{ref}}$ is the reference (correct) answer, $S$ is the reference solution, $A_{\text{stud}}$ is the student's provided answer, and $I$ is an image of the student's scratchwork.

Our dataset is specifically structured to support two critical tasks:
\textbf{Error Cause Explanation (ECE)}: An open-ended explanation describing the specific reason for the student's error, denoted as $E$.
\textbf{Error Cause Classification (ECC)}: A categorical classification identifying the type of error from a predefined taxonomy, denoted as $C$.

\paragraph{Error Cause Taxonomy}

\begin{figure}[t]
    \centering
    \begin{minipage}[c]{0.41\textwidth}
        \captionof{table}{Dataset statistics for the \dataset}
        \label{tab:token_stats_updated}
        \centering
        \setlength{\tabcolsep}{4pt}
        \renewcommand{\arraystretch}{1}
        \resizebox{\linewidth}{!}{%
            \begin{tabular}{@{}lcc@{}}
            \toprule
            \textbf{Metric} & \textbf{Primary} & \textbf{Middle} \\
            \midrule
            Total Samples & 1479 & 241  \\
            \midrule
            Grade Distribution &
            \makecell[tl]{ 
                1-2: 23.0\% \\
                3-4: 46.3\% \\
                5-6: 30.7\%
            } &
            \makecell[tl]{ 
                7: 31.5\% \\
                8: 33.6\% \\
                9: 34.9\%
            } \\
            \midrule
            \multicolumn{3}{c}{\textit{Problem $Q$ and Solution $S$}} \\
            \midrule[0.02em]
            Unique Problem          & 1279 & 241  \\
            Avg $Q$ Token           & 61.1 & 61.5 \\
            Avg $S$ Token           & 139.4 & 175.6 \\
            \midrule
            \multicolumn{3}{c}{\textit{Error Cause Explanation $E$}} \\
            \midrule[0.02em]
            Avg $E$ Token           & 53.4 & 48.4 \\
            Min $E$ Token           & 4    & 26   \\
            Max $E$ Token           & 162  & 116  \\
            \bottomrule
            \end{tabular}
        }
    \end{minipage}
    \hfill 
    \begin{minipage}[c]{0.55\textwidth}
        \centering
        \includegraphics[width=\linewidth]{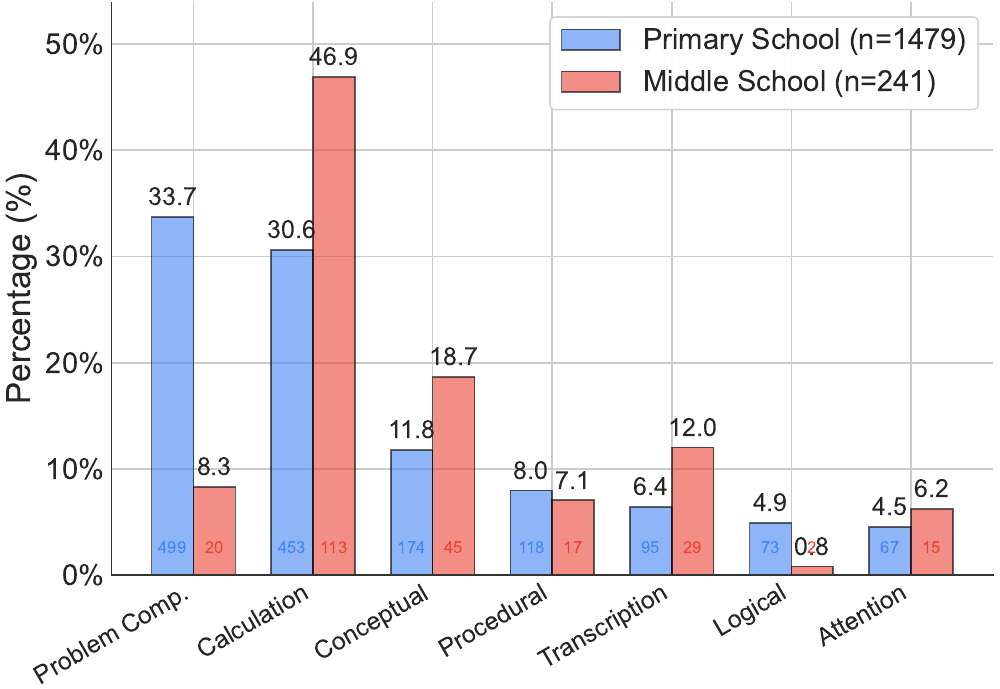}
        \caption{Distribution of error cause classification labels for primary school and middle school problems}
        \label{fig:error_type_distribution}
    \end{minipage}
\end{figure}

The taxonomy was systematically constructed through iterative expert reviews and educational theory-driven analysis on a much larger corpus of educational data, resulting in seven distinct error categories: \textit{Procedural Error, Calculation Error, Logical Reasoning Error, Transcription Error, Problem Comprehension Error, Conceptual Knowledge Error, and Attention and Detail Error}, with the quantity distribution shown in Figure~\ref{fig:error_type_distribution}. Notably, all error types are represented across both primary and middle school problems in our dataset.

\paragraph{Evaluation Metrics}
We employ two evaluation approaches aligned with the dual outputs required:
\textbf{Error Cause Explanation (ECE)}: We use an LLM-as-a-Judge framework, which assesses the semantic alignment of model-generated explanations with ground truth.
\textbf{Error Cause Classification (ECC)}: Error classification is evaluated using accuracy (Acc), strictly considering correct only those cases where the predicted class exactly matches the annotated class. This strict criterion emphasizes precision in classification performance.

\begin{figure*}[t]
    \centering
    \includegraphics[width=1\linewidth]{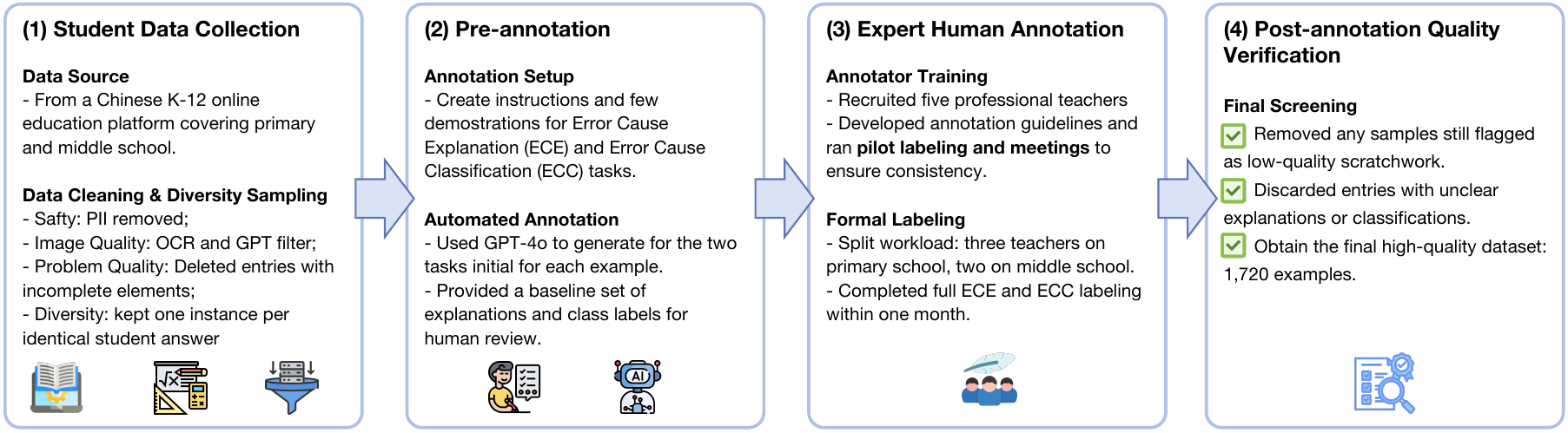}
    \caption{An overview of the \dataset benchmark construction pipeline.}
    \label{fig:data_construction}
\end{figure*}

\subsection{Dataset Construction}

Recent studies have shown that data contamination is a prevalent concern in MLLM benchmarks~\cite{song2024leaked}; our dataset mitigates this risk as it consists entirely of original, unpublished student scratchwork. As shown in Figure~\ref{fig:data_construction}, the dataset construction consists of four parts.

\paragraph{Part I. Student Data Collection}

\textbf{Data Source.} Student data were sampled from an online education platform, covering primary (grades 1-6) and middle school (grades 7-9) math questions. Students completed teacher-assigned tasks and received feedback. \textbf{Data Cleaning.} For the data safety, sensitive personally identifiable information (PII) was removed, retaining only relevant content related to answering questions. We also applied dual filtering using OCR tools and the GPT-4o-mini model to remove low-quality scratchwork images, such as illegible text or significant blurring. Entries with incomplete questions or missing answers were deleted, and text formatting was corrected. Questions containing images were excluded to simplify the initial recognition task. \textbf{Diversity Sampling.} To maintain diversity, only one instance per identical student answer for the same question was retained, resulting in around 3,400 distinct questions from an initial pool of about 1.1 million entries.

\paragraph{Part II. Answer Pre-annotation}

To reduce human workload and accelerate annotation efficiency, we adopted a human-computer collaborative approach for data annotation, referencing methods from previous works \cite{zhang2024mathverse,zhou2024mathscape}. Leveraging the gpt-4o-2024-05-13 model, known for its robust performance in generating preliminary annotations, we created initial Error Cause Explanation and Error Cause Classification responses for each question.

\paragraph{Part III. Expert Human Annotation}

Our labeling pipeline combined advanced automated methods with expert human validation to ensure high annotation quality. 
We engaged five professional mathematics teachers based in Beijing, each possessing over three years of teaching experience at primary and middle school levels. Teachers were remunerated at a rate of at least 60 RMB per hour. The annotation workload was strategically divided, with three teachers focusing on primary-level questions and two dedicated to middle-school-level queries.
The annotation procedure was systematically structured into three core stages:

\textit{Stage 1: Human Annotation Training.} Annotators were extensively trained by the researchers to revise and validate GPT-generated annotations. Training sessions included detailed guidance and annotation rules clearly articulated through example image prompts. 

\textit{Stage 2: Trial Annotation.} Moreover, annotators underwent trial annotation sessions using a standardized set of 30 questions. Post-session discussions facilitated clarification, resolution of uncertainties, and refinement of annotation guidelines, a process that was iterated until an inter-annotator agreement (IAA) of over 90\% was achieved by the annotators on this standardized set, ultimately ensuring consistency and accuracy in labeling.

\textit{Stage 3: Formal Annotation.} Following two comprehensive team meetings to finalize annotation protocols, annotators commenced formal labeling. The annotation process for all 3,400 questions was completed within one month.

\paragraph{Part IV. Post-annotation Quality Verification}
To further enhance dataset quality, post-annotation verification involved two additional screening phases. Firstly, scratchwork entries identified by annotators as low-quality were eliminated. Secondly, entries where the error cause or classification was indeterminate were discarded, culminating in the final, high-quality dataset with 1,720 entries.

\subsection{Data Statistics}

Our dataset includes 1,720 math problems, spanning 1,479 primary and 241 middle school problems, carefully selected to ensure rich coverage and representativeness. Detailed statistics, including precise grade distributions and comprehensive token counts for questions, solutions, and error explanations, are concisely presented in Table \ref{tab:token_stats_updated}.
A notable diversity is evident in error distributions across educational levels (see Figure \ref{fig:error_type_distribution}), highlighting distinct challenges students encounter at different stages of learning.
To ensure educational relevance and alignment, mathematical topics are categorized according to the authoritative Chinese \textit{Compulsory Education Curriculum Plan and Standards (2022 edition)} (Table \ref{tab:knowledge_points_distribution}). Primary-level questions predominantly address foundational areas such as numbers, expressions, geometry, and applied mathematics, while middle-school-level problems delve deeper into equations, functions, and advanced algebraic concepts.

\begin{table*}[t]
\caption{Distribution and Examples of Mathematical Topics by Category in \dataset (translated in English)}
\label{tab:knowledge_points_distribution}
\centering
\renewcommand{\arraystretch}{1}
\setlength{\tabcolsep}{4pt}
\small
\resizebox{1\linewidth}{!}{%
\begin{tabular}{p{3.5cm}ccp{8.5cm}}
\toprule
\textbf{Category} & \textbf{Pri. (\%)} & \textbf{Mid. (\%)} & \textbf{Description and Examples} \\
\midrule
Numbers and Expressions & 45.8 & 30.7 & Arithmetic operations, fractions, decimals, algebraic identities. \textit{e.g., Multiples of 3, Simplified addition} \\
\midrule
Equations and Functions & 1.9 & 43.6 & Equations, inequalities, function analysis. \textit{e.g., Quadratic equations, Linear functions} \\
\midrule
Geometry and Measurement & 25.0 & 20.2 & Areas, perimeters, volumes, geometry. \textit{e.g., Isosceles trapezoid area, Pythagorean theorem} \\
\midrule
Applied Mathematics & 25.4 & 1.6 & Practical math, unit conversions, financial calculations. \textit{e.g., Cost calculations, Average speed} \\
\midrule
Statistics and Probability & 1.9 & 4.0 & Data collection, probability estimation. \textit{e.g., Statistical tables, Frequency probability} \\
\bottomrule
\end{tabular}
}
\end{table*}

\section{Experiments}

\subsection{Experiment Setup}

\paragraph{Evaluated Models}

We selected 16 representative MLLMs for benchmarking on our dataset, covering a wide spectrum of model sizes and architectures. The evaluated models include 10 open-source models: 
Qwen2.5-VL (7B, 72B) \cite{bai2025qwen2}, DeepSeek-VL2 \cite{wu2024deepseek}, Phi-4-Multimodal \cite{abouelenin2025phi}, Llama-3.2-Vision (11B, 90B) \cite{grattafiori2024llama}, Gemma-3 \cite{team2025gemma}, Skywork-R1V \cite{peng2025skywork}, QVQ \cite{qvq-72b-preview}, and InternVL2.5 \cite{chen2024expanding}; 
as well as 6 proprietary models: Gemini 2.0 Flash (Flash-Lite, Flash Thinking) \cite{team2023gemini}, GPT-4o (GPT-4o mini, o4-mini\footnote{https://openai.com/index/introducing-o3-and-o4-mini/}) \cite{hurst2024gpt}.

\paragraph{Prompting and Hyperparameters}
For consistency and fairness in comparison, we standardized the prompting approach across all tested models. Specifically, we utilized a structured prompt during testing.
To further assess prompting effects, we conducted additional Chain-of-Thought (CoT) experiments, revealing some improvements in ECC task performance.
To ensure reproducibility and comparability, we set the generation temperature to 0 (greedy decoding), the maximum output length to 2048 tokens, and evaluated open-source models using NVIDIA A800-80G GPUs.
For proprietary reasoning models, we adopted their recommended default temperature settings, and additional experiments confirmed minor performance fluctuations when employing higher temperature settings.

\begin{table*}[t]
\caption{\textbf{Performance of state-of-the-art MLLMs on \dataset}.
We use weighted-average accuracy for the ECC task.
\textbf{Boldface} and \underline{underline} mark the best and second-best results for each metric, reported \textit{separately} for proprietary vs.\ open-source groups.
$^\dag$: activated parameters of MoE model.
$^*$ indicates \emph{reasoning} models. }
\label{tab:main_result}
\centering
\renewcommand{\arraystretch}{1.1}
\resizebox{1\textwidth}{!}{%
\begin{tabular}{l|c|ccc|ccc|cc}
\toprule
\multirow{2}{*}{Model} & \multirow{2}{*}{\#Params} &
\multicolumn{3}{c|}{Error Cause Explanation} &
\multicolumn{3}{c|}{Error Cause Classification} &
\multirow{2}{*}{Average} & \multirow{2}{*}{Avg Rank}\\
 & & \textbf{Primary} & \textbf{Middle} & \textbf{Rank} &
     \textbf{Primary} & \textbf{Middle} & \textbf{Rank} & & \\ 
\midrule[0.12em]
{\itshape Random Guessing} & {\itshape --} &
{\itshape 0.0}  & {\itshape 0.0}  & {\itshape --} &
{\itshape 12.5} & {\itshape 12.5} & {\itshape --} &
{\itshape 6.25} & {\itshape --} \\
{\itshape Human Performance} & {\itshape --} &
{\itshape 89.3} & {\itshape 86.2} & {\itshape --} &
{\itshape 78.4} & {\itshape 81.5} & {\itshape --} &
{\itshape 83.9} & {\itshape --} \\
 \midrule[0.10em]
\multicolumn{10}{c}{\textbf{Proprietary Models}} \\ \midrule[0.06em]
Gemini 2.0 Flash           & --  & 52.2& 46.9& 4& 38.6& \textbf{49.0}& 2& 46.7& 3
\\
Gemini 2.0 Flash Lite      & --  & 36.0& 32.0& 8& 34.6& 46.5& 5& 37.2& 5
\\
Gemini 2.0 Flash Thinking$^*$  & --  & \underline{65.9}& \underline{61.0}& 2& \textbf{43.9}& \underline{47.3}& 1& 54.5& 2
\\
GPT\textendash4o           & --  & 47.7& 44.8& 5& 26.1& 22.0& 11& 35.2& 9
\\
GPT\textendash4o mini      & --  & 3.5& 1.2& 16& 20.1& 14.1& 13& 9.8& 16
\\
o4-mini$^*$   & --  & \textbf{71.8}& \textbf{69.7}& 1& \underline{40.1}& \underline{47.3}& 3& 57.2& 1\\
\midrule[0.10em]

\multicolumn{10}{c}{\textbf{Open-Source ($<$ 10 B)}} \\ \midrule[0.06em]
Qwen2.5-VL                 & 7 B   & 15.6& 12.4& 15& 21.0& 11.6& 14& 15.2& 15
\\
DeepSeek-VL2               & 4.5 B$^\dag$ & 20.9& 25.7& 12& 16.6& 7.9& 16& 17.8& 14
\\
Phi-4-Multimodal           & 5.6 B$^\dag$ & 12.0& 18.3& 14& 28.9& 32.8& 9& 23.0& 12\\
\midrule[0.08em]

\multicolumn{10}{c}{\textbf{Open-Source (10-40 B)}} \\ \midrule[0.06em]
Llama-3.2-Vision           & 11 B & 13.4& 20.3& 13& 17.6& 26.1& 12& 19.4& 13
\\
Gemma-3                    & 27 B & 38.9& 26.1& 9& \underline{32.2}& \underline{46.1}& 6& 35.8& 7
\\
Skywork-R1V$^*$                & 38 B & 37.5& 33.6& 7& 27.7& 43.2& 8& 35.5& 8\\
\midrule[0.08em]

\multicolumn{10}{c}{\textbf{Open-Source ($>$ 40 B)}} \\ \midrule[0.06em]
QVQ$^*$       & 72 B & \textbf{57.5}& \textbf{56.8}& 3& 12.7& 17.0& 15& 36.0& 6
\\
Qwen2.5-VL                 & 72 B & \underline{40.0}& \underline{34.0}& 6& \textbf{32.5}& \textbf{49.4}& 4& 39.0& 4
\\
InternVL2.5                & 78 B & 27.1& 24.5& 11& 30.7& 44.8& 7& 31.8& 10
\\
Llama-3.2-Vision           & 90 B & 27.7& 26.1& 10& 15.9& 45.6& 10& 28.8& 11\\
\bottomrule
\end{tabular}} 
\end{table*}

\paragraph{Validation of LLM-as-a-judge}

As described in \S\ref{sec:dataset1}, we employed the LLM-as-a-judge metric to evaluate the Error Cause Explanation (ECE) task. To validate its reliability, we conducted an experiment using 70 randomly sampled ECE cases, evaluated by the advanced LLM, o3-mini. Manual verification showed the judge's accuracy reached 88.6\%, close to human-human inter-annotator agreement of 91.4\%, confirming its suitability for our evaluation.
The accuracy was below 100\% primarily because the judge occasionally identified plausible yet unannotated error reasons as mismatches.
We selected o3-mini due to its optimal trade-off between accuracy and evaluation cost, with the total cost of evaluating the entire benchmark (1,720 cases) being less than 10 USD.

\subsection{Main Results}

Table~\ref{tab:main_result} summarizes the performance of MLLMs on our benchmark. Key findings include:
\textbf{(1) Proprietary Models Outperform Open-source Models.}
Proprietary models consistently outperform open-source models even at similar parameter scales, likely benefiting from more diverse training data. However, a considerable gap remains compared to human performance, emphasizing the benchmark's inherent challenge.
\textbf{(2) Scaling Law on Both Tasks and Reasoning Model Superiority in ECE.}
Performance generally follows scaling laws, with larger models showing better results. Reasoning models specifically excel in Error Cause Explanation (ECE), highlighting their advantage in tasks demanding deeper semantic understanding. Conversely, Error Cause Classification (ECC) remains significantly more challenging across all models.
\textbf{(3) Elementary Tasks Not Necessarily Easier.}
While models typically perform better on primary tasks in the ECE task, primary-level performance unexpectedly falls below middle school performance in the ECC task. This could stem from less structured and harder-to-interpret handwriting in primary-level scratchwork, complicating precise error classification.

\section{Further Analysis}

We analyze three key research questions to deepen our understanding of model performance:

\noindent\textbf{RQ1:} What challenges do MLLMs face in error cause detection?\\
\noindent\textbf{RQ2:} How does problem type affect model performance?\\
\noindent\textbf{RQ3:} How does problem difficulty affect model performance?

\begin{table*}[t]
\caption{Examples (translated in English) illustrating typical model error types in ECE. ``Ref. Ans.'' and ``Stu. Ans.'' denote the reference and student answers, respectively. Bold text highlights the erroneous parts of the prediction.}
\label{tab:error_types_en}
\centering
\setlength{\tabcolsep}{4pt}      
\resizebox{1\textwidth}{!}{%
\renewcommand{\arraystretch}{1.3}
\begin{tabular}{C{3.5cm}L{3.7cm}C{1.2cm}C{1.3cm}L{3.8cm}L{5.5cm}}
\toprule
\textbf{Scratchwork} &
\textbf{Problem Statement} &
\textbf{Ref.}\newline\textbf{Ans.} &
\textbf{Stu.}\newline\textbf{Ans.} &
\textbf{Ground Truth} &
\textbf{Model Prediction} \\
\midrule

\multicolumn{6}{c}{\textit{\textbf{Visual Recognition Failure}}} \\
\cdashline{1-6}
\includegraphics[width=\linewidth,keepaspectratio,height=3.2cm]{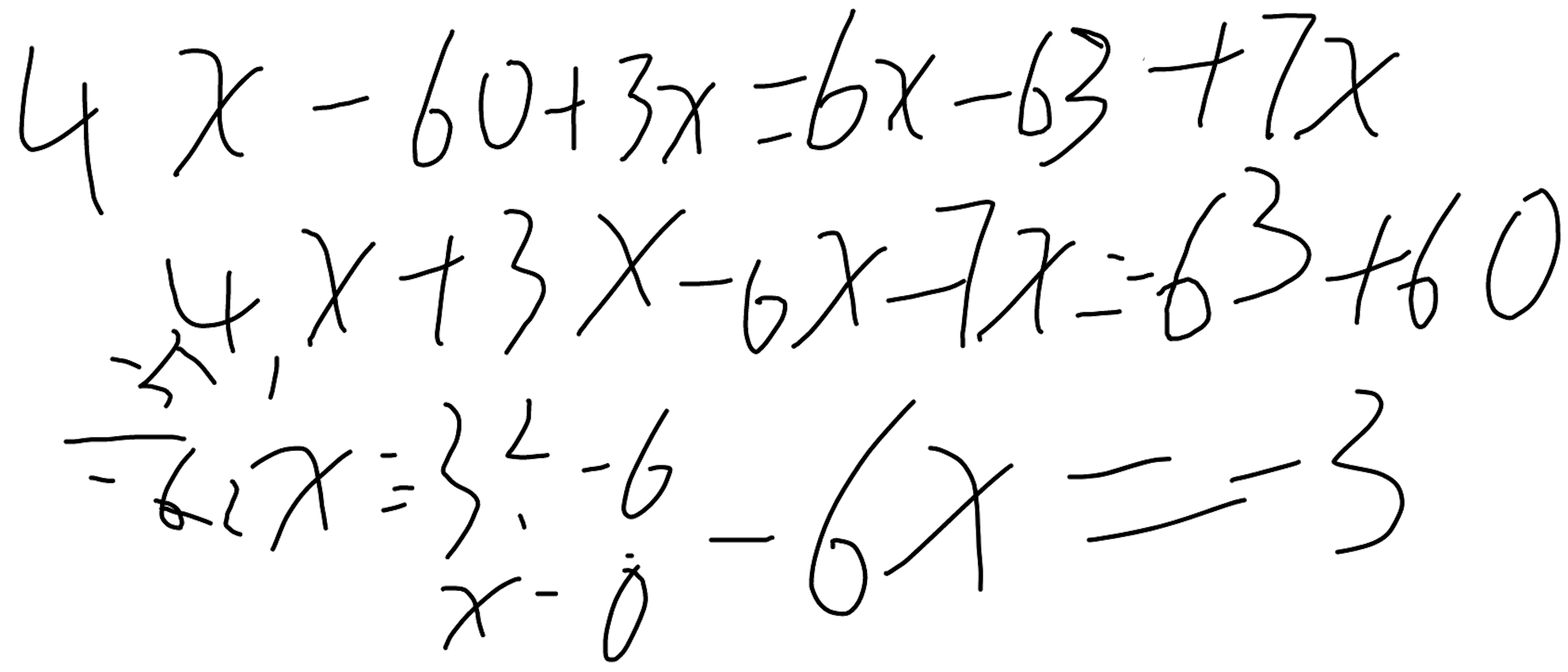} &
Solve for $x$: $4x-3(20-x)=6x-7(9-x)$ &
$\dfrac12$ &
$-\dfrac12$ &
Student solved $-6x=-3$ as $x=-\tfrac12$ instead of $\tfrac12$. &
...Finally $x = \frac{-3}{-6} = \frac{1}{2}$. However, \textbf{the student miscalculated $-63 + 60$ as $+3$ at this step, so they wrote $-6x = 3$}, leading to $x = \frac{3}{-6} = -\frac{1}{2}$.\\
\midrule

\multicolumn{6}{c}{\textbf{\textit{Formatting Misinterpretation}}} \\
\cdashline{1-6}
\includegraphics[width=\linewidth,keepaspectratio,height=3.2cm]{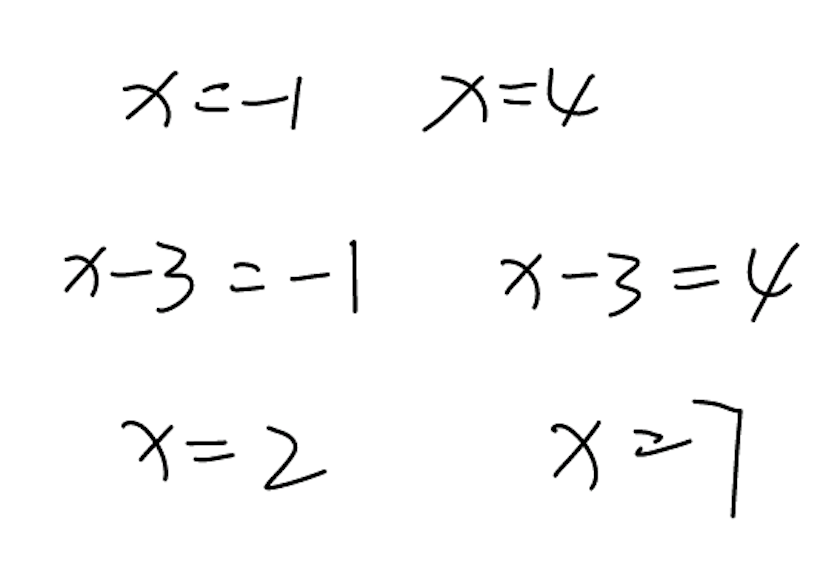} &
If the two roots of $a(x+m)^2+b=0$ are $-1$ and $4$, then the solutions to $a(x+m-3)^2+b=0$ are $x=$\underline{\hspace{1cm}}. &
\makecell{2 or 7} &
\makecell{2 and 7} &
Formatting error (“and” vs. “or”); also mis-shifted the quadratic by $m$. &
... The student, in their scratchpad, \textbf{mistakenly treated the second equation $a \cdot (x + m - 3)^2 + b = 0$ as $a \cdot (x - 3)^2 + b = 0$}. They then directly set $x - 3 = -1$ and $x - 3 = 4$, thus obtaining $x = 2, 7$... \\
\midrule

\multicolumn{6}{c}{\textbf{\textit{Misaligned Misinterpretation}}} \\
\cdashline{1-6}
\includegraphics[width=\linewidth,keepaspectratio,height=3.2cm]{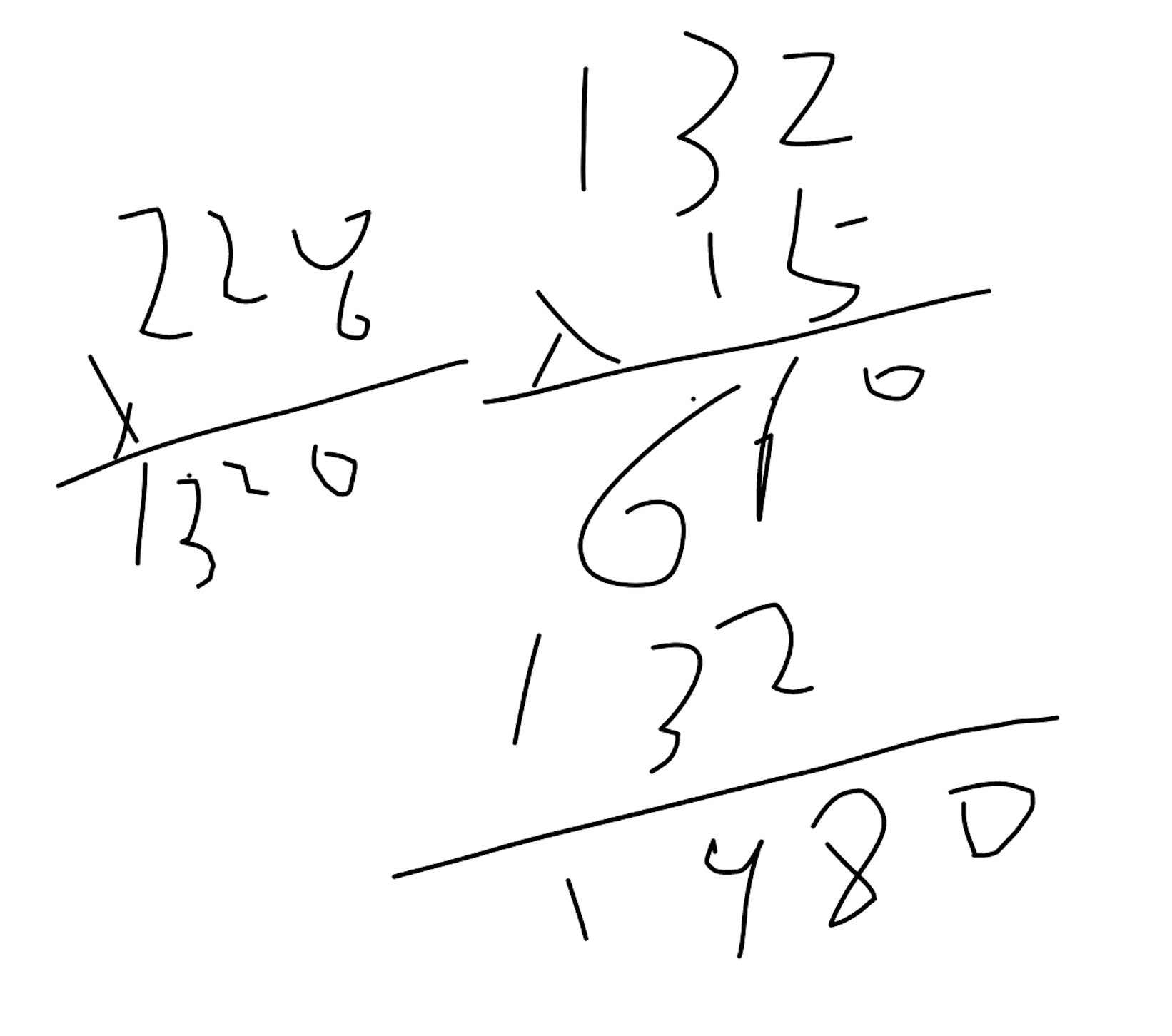} &
Brick dimensions $20\times11\times6$ cm, density $1.5$ g/cm$^{3}$.  
Find the weight in kilograms. &
1.98 &
1980 &
Answered in grams without converting to kg. &
...When calculating by hand, \textbf{he even converted "$1320 \times 1.5$" to "$132 \times 15$"}, and although the result was still $1980$, \textbf{the division by 10 and multiplication by 10 in the process didn't correspond to the units}... \\
\bottomrule
\end{tabular}}
\end{table*}

\subsection{RQ1: Challenges in Error Identification}

\paragraph{Qualitative Analysis}
We conducted detailed case studies to further illustrate typical errors made by the o4-mini. 
Table~\ref{tab:error_types_en} presents three representative cases categorized by the type of error: Visual Recognition Failure, Formatting Misinterpretation, and Misaligned Misinterpretation.
These examples highlight specific aspects that need improvement, such as visual processing accuracy, proper understanding of formatting requirements, and accurate inference of the student reasoning processes.

\paragraph{Quantitative Anlysis}
\begin{wrapfigure}{r}{0.45\textwidth} 
    \centering
    \includegraphics[width=0.42\textwidth]{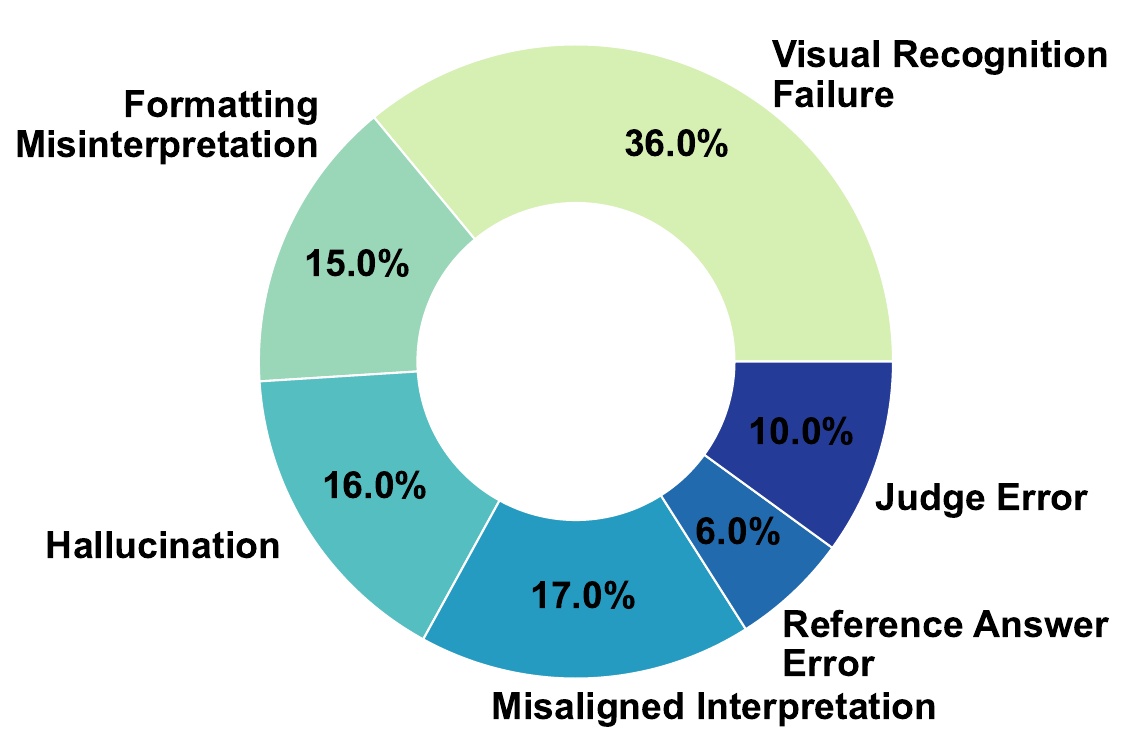}
    \caption{Failure Cases Distribution of o4-mini on ECE}
    \label{fig:error_distribution}
\end{wrapfigure}
To explore the difficulties faced by current MLLMs, we conducted an error analysis of 100 randomly selected cases in which the strongest model (o4-mini) failed on the Error Cause Explanation task. 
As shown in Figure~\ref{fig:error_distribution}, we categorize these errors into six types. 
Key findings include: 
(1) The most frequent errors were related to OCR and image recognition, often stemming from unclear handwriting.
(2) Models struggled significantly with accurately reconstructing students' reasoning processes, indicating limitations in logical inference.
(3) Many errors involved over-inference or speculative reasoning by the models, suggesting tendencies to extrapolate beyond available evidence.
In addition, to understand error patterns in smaller, open-source models, we conducted a similar analysis on Qwen2.5-VL-7B. We found a higher incidence of hallucination errors (22\%) and a new category, ``Model Calculation Error'' (17\%), indicating arithmetic reasoning difficulties specific to smaller models.

\subsection{RQ2: Impact of Problem Type on Performance}

\paragraph{Performance Across Error Cause Categories}

\begin{wrapfigure}{r}{0.57\textwidth} 
    \centering
    \vspace{-4mm}
    \includegraphics[width=0.55\textwidth]{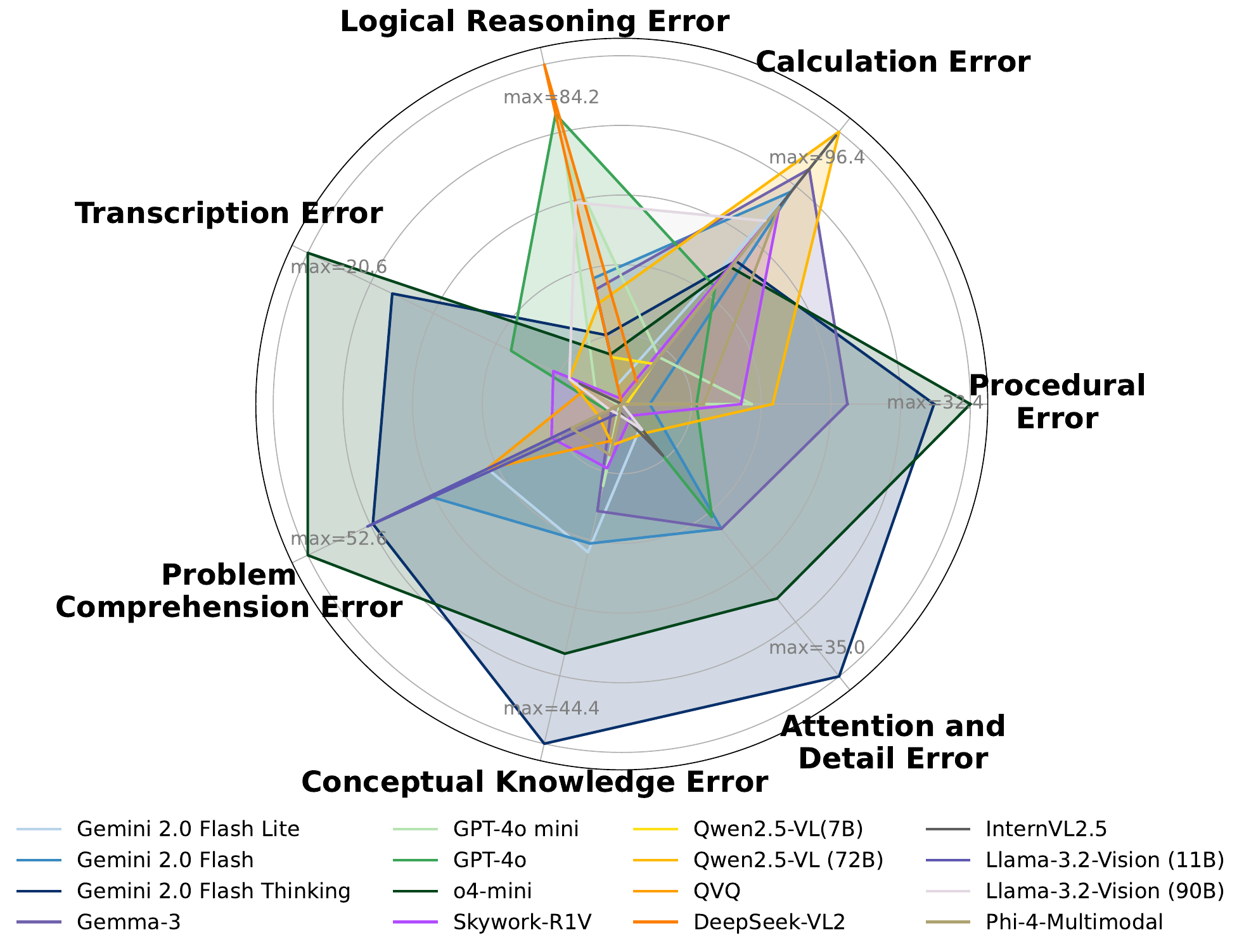}
    \caption{Models' Performance on different ECC classes}
    \label{fig:model_performance_radar}
\end{wrapfigure}
We investigated how problem categories influence model performance on ECC Task, averaging scores across primary and middle school datasets. 
Several insights emerged from the results shown in Figure~\ref{fig:model_performance_radar}: 
(1) The top-performing models, o4-mini and Gemini 2.0 Flash Thinking, notably excelled in most error categories, except Logical Reasoning and Calculation Errors, which are harder due to implicit reasoning steps and compounded errors in visual number recognition and multi-step arithmetic. 
(2) Many models demonstrated potential overfitting to specific error categories, particularly Logical Reasoning and Calculation Errors, indicating specialized rather than generalized error detection capabilities.
(3) Procedural and Transcription errors generally posed significant challenges to all models, highlighting areas for further targeted development.
Performance disparities across error types suggest varied levels of complexity inherent in different problem categories, reflecting a nuanced interaction between model architecture and problem characteristics.

\paragraph{Performance Across Mathematics Topics}

\begin{figure}[t]
    \centering
    \begin{minipage}{0.54\linewidth}
        \centering
        \includegraphics[width=\linewidth]{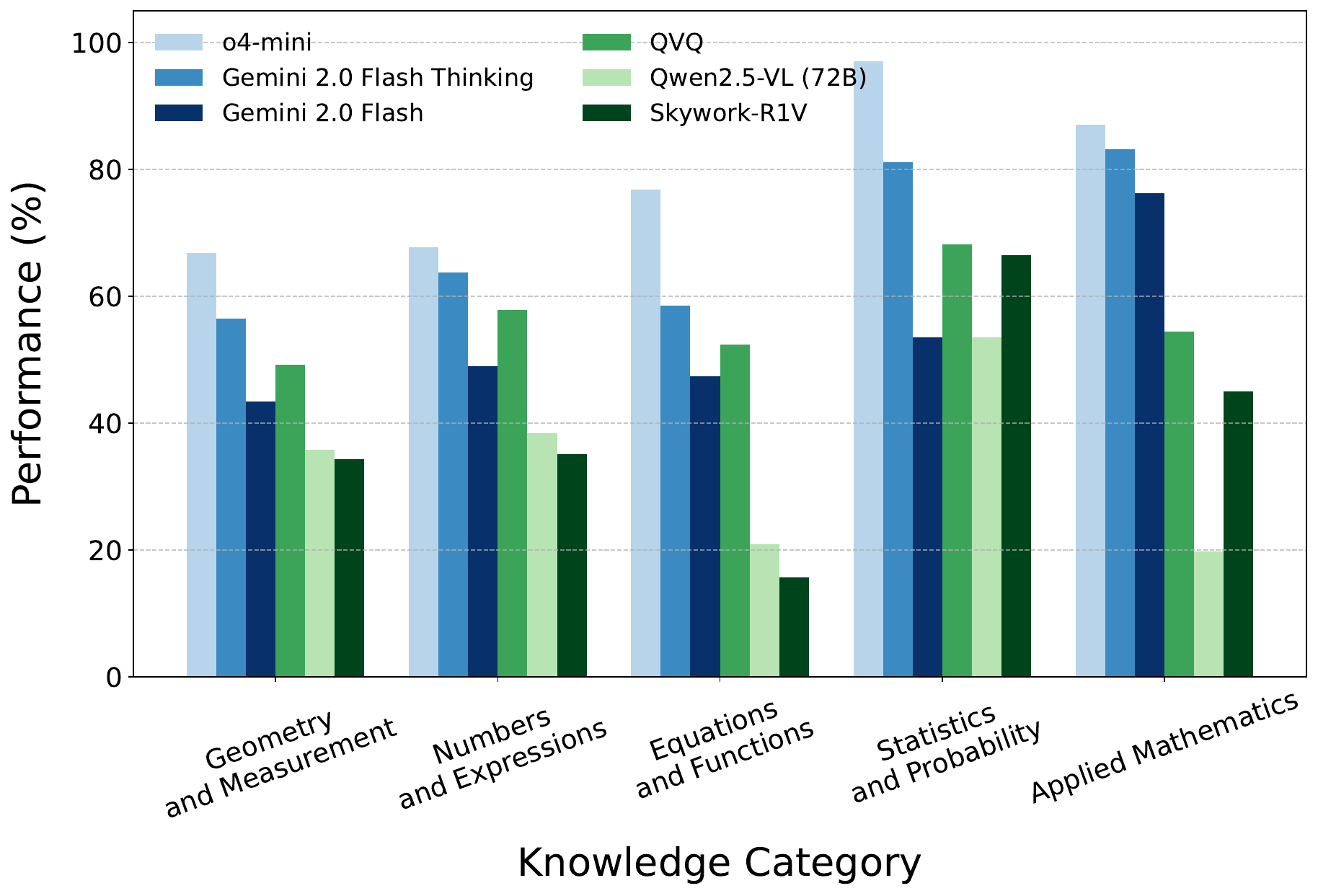}
        \caption{Model Performance Across Math Topics in ECE Tasks (Top 3 Open-source and Proprietary Models)}
        \label{fig:knowledge_category_performance}
    \end{minipage}
    \hfill
    \begin{minipage}{0.43\linewidth}
        \centering
        \includegraphics[width=\linewidth]{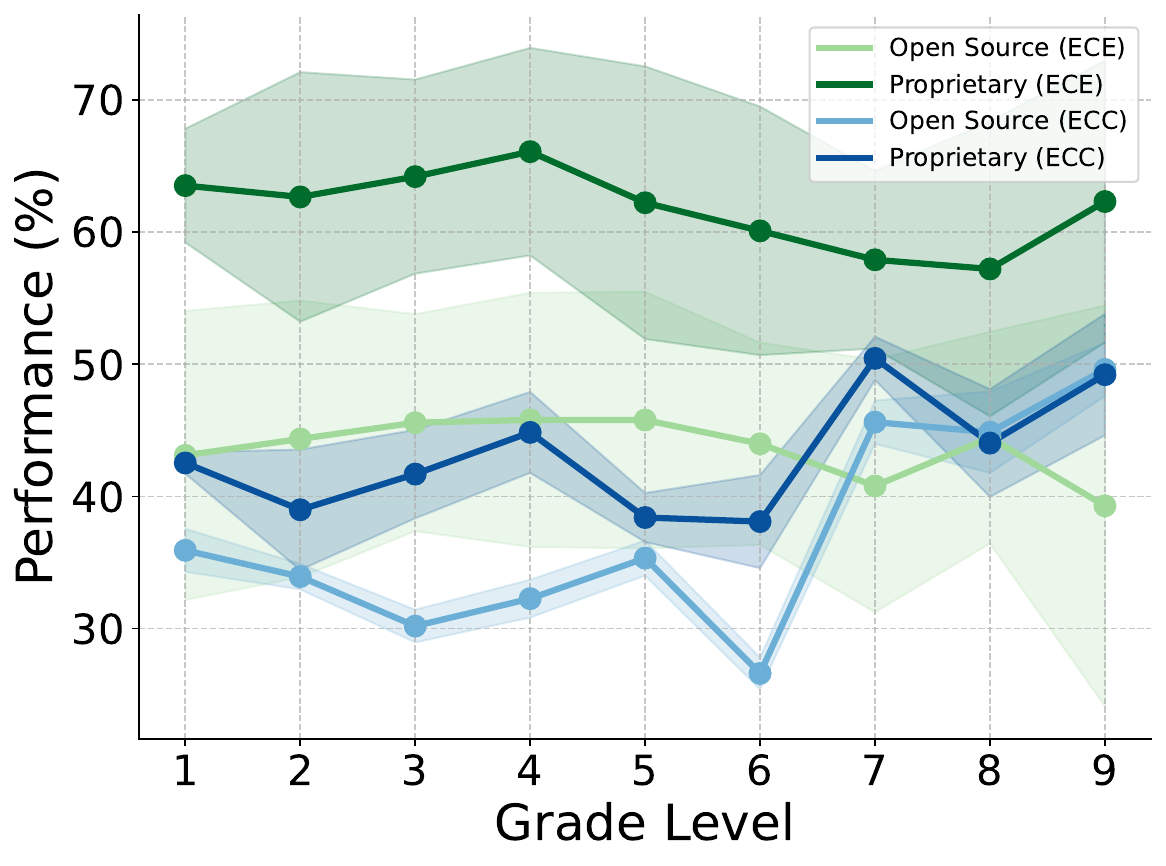}
        \caption{Performance across grade levels, averaged for top-3 open-source and proprietary models on two tasks.}
        \label{fig:model_performance_by_grade}
    \end{minipage}
\end{figure}

We also analyzed model performance based on the topics of math problems (introduced in Table~\ref{tab:knowledge_points_distribution}). 
Figure~\ref{fig:knowledge_category_performance} illustrates several notable findings:
(1) Proprietary models consistently showed strong and stable performance across all knowledge categories, with o4-mini significantly outperforming others.
(2) Open-source models exhibited varied performance, with Skywork-R1V notably stronger in Statistics and Probability and Applied Mathematics, yet weaker in Equations and Functions. The disparity in open-source model performance indicates a potential specialization or bias in training data, highlighting the importance of diverse and comprehensive training datasets.

\subsection{RQ3: Impact of Difficulty on Model Performance}

Additionally, we examined the impact of educational grade level on model performance by selecting the top three open-source and proprietary models from ECC and ECE tasks.
The analysis, depicted in Figure~\ref{fig:model_performance_by_grade}, suggests several trends:
(1) On the ECE task, model performance exhibits a slight downward trend as grade level increases, indicating greater complexity or ambiguity in higher-grade solutions.
(2) Conversely, performance on the ECC task generally improves with increasing grade levels, possibly due to clearer and more structured scratchwork provided by older students. Through detailed sampling, we observed that middle-school scratchwork contains clearer sequential steps and standardized notation compared to elementary-level responses. These structured presentations facilitate error classification.
(3) Proprietary models consistently outperform open-source models across all grades, underscoring the potential advantages of more diverse training data.

\section{Conclusion and Future Work}
In summary, \dataset advances educational AI by introducing a comprehensive multimodal benchmark that exposes the limitations of current MLLMs in diagnosing student errors. It highlights the urgent need for developing models that better align with educators' analytical processes.
A limitation of this work is that all samples were collected from Chinese students using a single online education platform, which may constrain generalizability to other languages, demographics, and educational contexts.
Future work includes enhancing model training by incorporating explicit error-type predictions, integrating advanced visual recognition techniques, and exploring step-by-step reasoning alignment to improve model interpretability. Expanding the dataset across diverse populations and educational settings, and leveraging cross-cultural comparisons, could yield deeper insights into universally effective educational strategies.

\begin{credits}
\subsubsection{\discintname} The authors have no competing interests to declare that are relevant to the content of this article.
\end{credits}

\bibliographystyle{splncs04}
\bibliography{mybibliography}

\end{document}